\documentclass[conference]{IEEEtran}
\IEEEoverridecommandlockouts
\makeatletter
\usepackage{cite}
\makeatother
\usepackage[colorinlistoftodos]{todonotes}
\usepackage{amsmath}
\usepackage{amsmath,amssymb,amsfonts}
\usepackage{algorithmic}
\usepackage{graphicx}
\usepackage{textcomp}
\usepackage{xcolor}
\usepackage{ctable}
\def\BibTeX{{\rm B\kern-.05em{\sc i\kern-.025em b}\kern-.08em
   T\kern-.1667em\lower.7ex\hbox{E}\kern-.125emX}}
\usepackage{amsmath,amssymb,amsfonts}
\usepackage{algorithmic}
\usepackage{graphicx}
\usepackage{textcomp}
\usepackage{url}

\usepackage[breaklinks]{hyperref}
\usepackage{graphicx}
\usepackage{booktabs}
\usepackage[flushleft]{threeparttable}
\usepackage{float}
\usepackage{nth}
\usepackage{siunitx}
\usepackage{afterpage}
\usepackage{cite}
\usepackage{dblfloatfix}
\usepackage{array,booktabs}
\usepackage{multicol}
\ifCLASSOPTIONcompsoc \usepackage[caption=false,font=normalsize,labelfon t=sf,textfont=sf]{subfig} \else \usepackage[caption=false,font=footnotesize]{subfig} \fi

\begin{document}

\title{Privacy-Preserving Eye-tracking Using Deep Learning*\\
\thanks{ This research was supported by the National Science Foundation, grant \# 1822378,  
 the NIH under grant \# UL1TR002378 and P30AG066511, the
 James M. Cox Foundation and Cox Enterprises Inc., and the Goizueta Foundation.}
}

\author{\IEEEauthorblockN{Salman Seyedi}
\IEEEauthorblockA{dept. of Biomedical Informatics\\ Emory School of Medicine\\Atlanta, Georgia\\sseyedi@emory.edu}
\and
\IEEEauthorblockN{Zifan Jiang}
\IEEEauthorblockA{dept. of Biomedical Engineering\\ Georgia Institute of Technology\\Atlanta, Georgia\\zifanjiang@gatech.edu}
\and
\IEEEauthorblockN{Allan Levey}
\IEEEauthorblockA{dept. of Neurology\\ 
Emory School of Medicine\\ Atlanta, Georgia\\alevey@emory.edu}
\and
\IEEEauthorblockN{Gari D. Clifford}
\IEEEauthorblockA{dept. of Biomedical Informatics\\ Emory School of Medicine\\ Atlanta, Georgia\\gari@gtech.edu}
}


\maketitle

\begin{abstract}
The expanding usage of complex machine learning methods like deep learning has led to an explosion in human activity recognition, particularly applied to health. 
In particular, as part of a larger body sensor network system, face and full-body analysis is becoming increasingly common for evaluating health status.
However, complex models which 
handle private and sometimes protected data, raise concerns about the potential leak of identifiable data. In this work, we focus on the case of a deep network model trained on images of individual faces. 

A previously published deep learning model, trained to estimate the gaze from full-face image sequences was stress tested for personal information leakage by a white box inference attack.
Full-face video recordings taken from 493 individuals undergoing an eye-tracking based evaluation of neurological function were used.
Outputs, gradients, intermediate layer outputs, loss, and labels were used as inputs for a deep network with an added support vector machine emission layer to recognize membership in the training data. 

The inference attack method and associated mathematical analysis indicate that there is a low likelihood of unintended memorization of facial features in the deep learning model. In this study, it is showed that the named model preserves the integrity of training data with reasonable confidence. The same process can be implemented in similar conditions for different models.



\end{abstract}

\section{Introduction} 
The importance of exploring guidelines and regulations regarding implementation for different machine learning (ML) and artificial intelligence (AI) techniques increases as these techniques become prevalent in various settings that involve private individual data.
In the US, in the context of health-related data and protected health information (PHI), the Health Insurance Portability and Accountability Act of 1996 (HIPAA) defines the restrictions on which information must be scrubbed prior to use outside of a protected enclave. HIPAA's primary goals are providing regulation to facilitate the portability of the data and preventing leakage of PHI. One of the main concerns in the increased use of deep learning models in the different private data or PHI is that these models, with an extensive number of variables and parameters, have the potential of encoding personal details\cite{melis2019exploiting} and, when shared, can result in an unintended data leak\cite{vepakomma2018no}. These vulnerabilities seem to be exploitable not only by black box attacks\cite{bhagoji2018practical} using only the outputs of models, but through the calculation of the gradients, loss, and other derivable parameters of the model and different inputs\cite{zhu2020deep}\cite{nasr2019comprehensive}.

Convolutional neural networks (CNN) can be particularly complex. The increased adoption of CNNs in the context of facial analysis and medical imaging \cite{zifan2021}\cite{li2018deep} raises concerns over their ability to encode private data.  
This work, therefore, explores a CNN-based model to stress-test under inference attacks, developed for an eye-tracking task
\cite{haque2019vismet}. 
This eye-tracking model can be divided into three parts. The first part involves a regression tree for face and eye detection. This detects the face and eyes from each frame in a recording. The second part, which is CNN-based and is the core of the pipeline, consists of three CNNs, one for each eye and one for the face, followed by a fully connected neural network (FCN) for eyes, face, and face grid. Then the outputs of three FCNs come as inputs to another FCN to estimate the eye gaze relative to the camera position. The third part involves a support vector regression over each recording to enhance the accuracy in the eye-tracking model, but is not included in this study, since it compresses inputs into two numbers (coordinates on a screen) and has little potential for encoding individual information. The main potential vulnerability lies in the CNN component of the system, where the face and eyes are processed by a large number of weights, and could therefore have the potential to memorize the facial features of the participants. More detail of the target model can be found in Haque {\em et al.} \cite{haque2020deep}

The key contributions in this paper are 1) The formulation of the privacy attack model, and 2) the demonstration that algorithms that analyze aspects of the human face that are not specific to any individual (at least with the complexity observed in our real-world mode) are unlikely to leak PHI.

\section{Materials and methods}

\subsection{Methodology}
The primary approach of this study in investigating the potential memorization of facial information in an eye-tracking model (herein referred to as the ``target model'') in two general aspects. 
The first aspect is the analysis of pipeline performance over the membership inference of recordings. In this approach, the overall success of the attacking pipeline in differentiating between recordings used in the target model training and the recordings that were not would be a metric for the amount of data memorization in the target model.

The second aspect is the further analysis of people with multiple recordings where one recording has been used to train the target model, and the other has not. These cases are of particular interest since the same face has been used in the target training but not the same recording. So, any boost in performance of the attacking pipeline for these cases would be indicative of the memory of facial information in the target model. 

\subsection{Dataset}
The dataset used in this work was made of 610 video recordings from 493 participants in the Emory Healthy Aging Study undergoing an eye-tracking based evaluation of neurological function \cite{haque2019vismet}\cite{haque2020deep}. Our goal here is to create an attack model on the previously designed eye-tracking model. Note that we have all the information about the training set of the target at this point in the study. In our attack model, recordings of participants with single recording were randomly divided into three separate sets: training, validation, and test sets. There were 54 participants with multiple recordings where they have at least one recording inside the training set of the target model and at least one recording not in the training set of the target model. The recordings from these 54 participants were divided into in\_training (those recordings used in the target model's training set) and out\_training (those recordings not in the target model's training set). The in\_training recordings were all put in the attack model training set as well. The out\_training recordings were randomly divided into training, validation, and test sets for the attack model ( Table \ref{table:Data}).

For the labeling for the attack model(s), (Y), two of them were produced. In $Y_{instance}$, the labels were set to (1) for all the frames if the recording was in the target network's training set and (0) otherwise. In $Y_{person}$, labels were set to 1 if at least one recording of the person was used to train the target network. In other words if there are two recordings of person A, A1 and A2, then if A1 was used in the training of target network but not A2, then $Y_{instance}(A1)=1$, $Y_{instance}(A2)=0$, but $Y_{person}(A1)=1$ and $Y_{person}(A2)=1$.


\begin{table}[htbp]
\caption{Data distribution}
\label{table:Data}
\begin{center}
\begin{tabular}{ll|l|l|l|}
\cline{3-5}
                                                         &                   & train & valid & test  \\ \hline
\multicolumn{1}{|l|}{number of Records} & total             & 242   & 170   & 198   \\ \cline{2-5}
\multicolumn{1}{|l|}{}                                   & Y$_{instance}$=1 & 140   & 73    & 93   \\ \cline{2-5} 
\multicolumn{1}{|l|}{}                                   & Y$_{person}$=1   & 159   & 87   & 114   \\ \hline
\multicolumn{1}{|l|}{number of Frames}                   & total             & 83477 & 66755 & 73857 \\ \cline{2-5} 
\multicolumn{1}{|l|}{}                                   & Y$_{instance}$=1 & 46515 & 30072 & 35999 \\ \cline{2-5} 
\multicolumn{1}{|l|}{}                                   & Y$_{person}$=1 & 50654 & 32950 & 41456 \\ \hline
\end{tabular}
\end{center}
\end{table}

\subsection{Classification Pipeline}
The pipeline can be divided into three parts, parameters collection, classifier/frame labeling, and patient membership inference.

\subsubsection{Parameters Collection}
The original trained eye-tracking model was used to get not only the activations, output, and label but also the gradients and loss for each frame in each recording. This was performed by feeding the trained network the frame and label and extracting the calculated parameters. 

\subsubsection{Classifier/Frame Labeling}
This part can be viewed as a two-step section, encoding and classifying. In the encoding part, for any frame, the parameters from the previous step were fed to a separate, FCN with one hidden layer, so the information gets encoded with specific encoders for each parameter type. For each input, they get encoded to a 64 dimension. Then the outputs of the encoding parts get fed to another FCN with three hidden layers to train for classification using encoded information (Moderately similar to the work by M.  Nasr, R. Shokri, and A. Houmansadr  \cite{nasr2019comprehensive}). The output of this second part is a number between 1 and 0 which is the probability the model assigns to the frame (see labels generation in experiment subsection) being in the training set for the original eye-tracking model or not. All activation functions are ReLU (Rectified Linear Unit), except the last one, which is the Sigmoid function, to produce the probabilities. Binary cross-entropy is used as the loss function for the training of the labeling network as a whole. 
\subsubsection{Patient Membership Inference}
The outputs of the classifier part are for each frame. However, any recording either has been part of the training the eye-tracking model or not. In this step, the labels of all the frames from each recording (the number of frames is different for different recordings) are used to produce a final membership inference for each recording. Here different moments (mean, variance, skewness, and kurtosis) and the entropy have been captured for each recording to train the support vector machine (SVM) to label each record. 
\subsection{Experiments}
Different sets of parameters have been used as input for the frame labeling part to find the best network based on the performance on the validation set. Also, in the patient membership inference section, some other models were investigated, but SVM with linear kernel provided the best results on the validation set, and so it was picked. All the steps in the pipeline have been done for two models, the instance model (trained on the data set with instance label) and the person model (trained on data set with person label). Fifty-four records have different Instance labels and Person labels. They have been assigned to the training, validation, and test sets (19 records have been assigned to the training set, 14 records to the validation set, and 21 to the test set). 
\section{Results}
A different set of parameters collected on step one of the classification pipeline was used to determine which ones provided information and improved the model results in step two. The loss (binary cross-entropy) on validation dataset for several sets can be seen in table \ref{table:loss_param}. Adding loss and label (from target model) with last layers gradients show improvement in the loss, and so the model with two outputs, three gradients, and loss (from target model) and label as input was selected for the rest of the work since it has the best performance. The value of binary cross-entropy (loss) for both models on test dataset is 0.59 ($loss^{test}_{Instance}=loss^{test}_{Person}=0.59$).

\begin{figure}[htbp]
\centering
\includegraphics[width=0.5\textwidth]{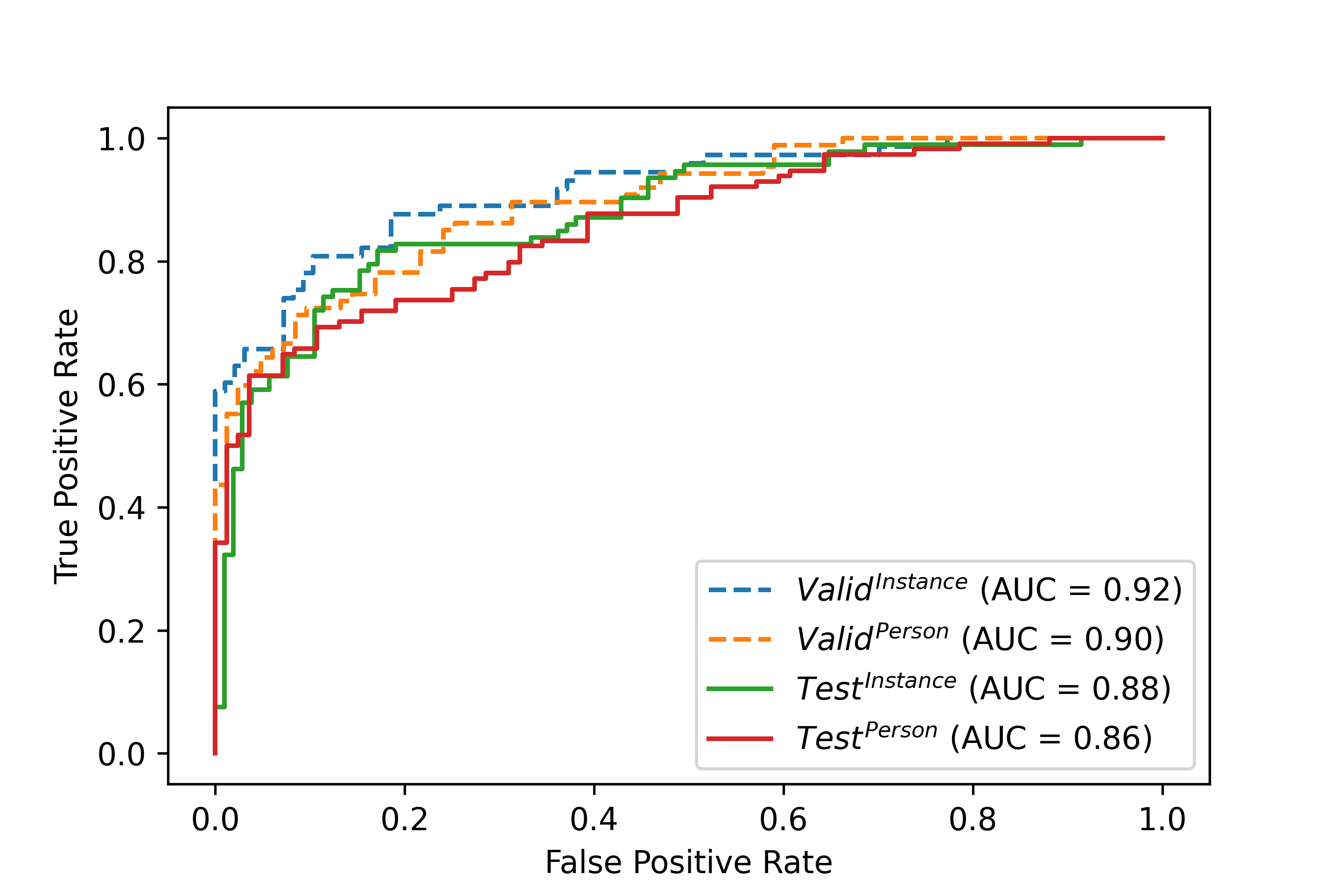}
    \caption{ROC curve (SVM on labeling video recordings): the dash-lines correspond to the validation set while the solid lines are for the test set. The area under the curve for all sets and labels has been shown in the legend. While the blue and green are for the dataset with instance labeling, the orange and red indicate values for the dataset with person labeling.}
    \label{fig:ROC}
\end{figure}

\begin{figure}[htbp]
\centering
\vspace{-0.4cm}
\includegraphics[width=0.5\textwidth]{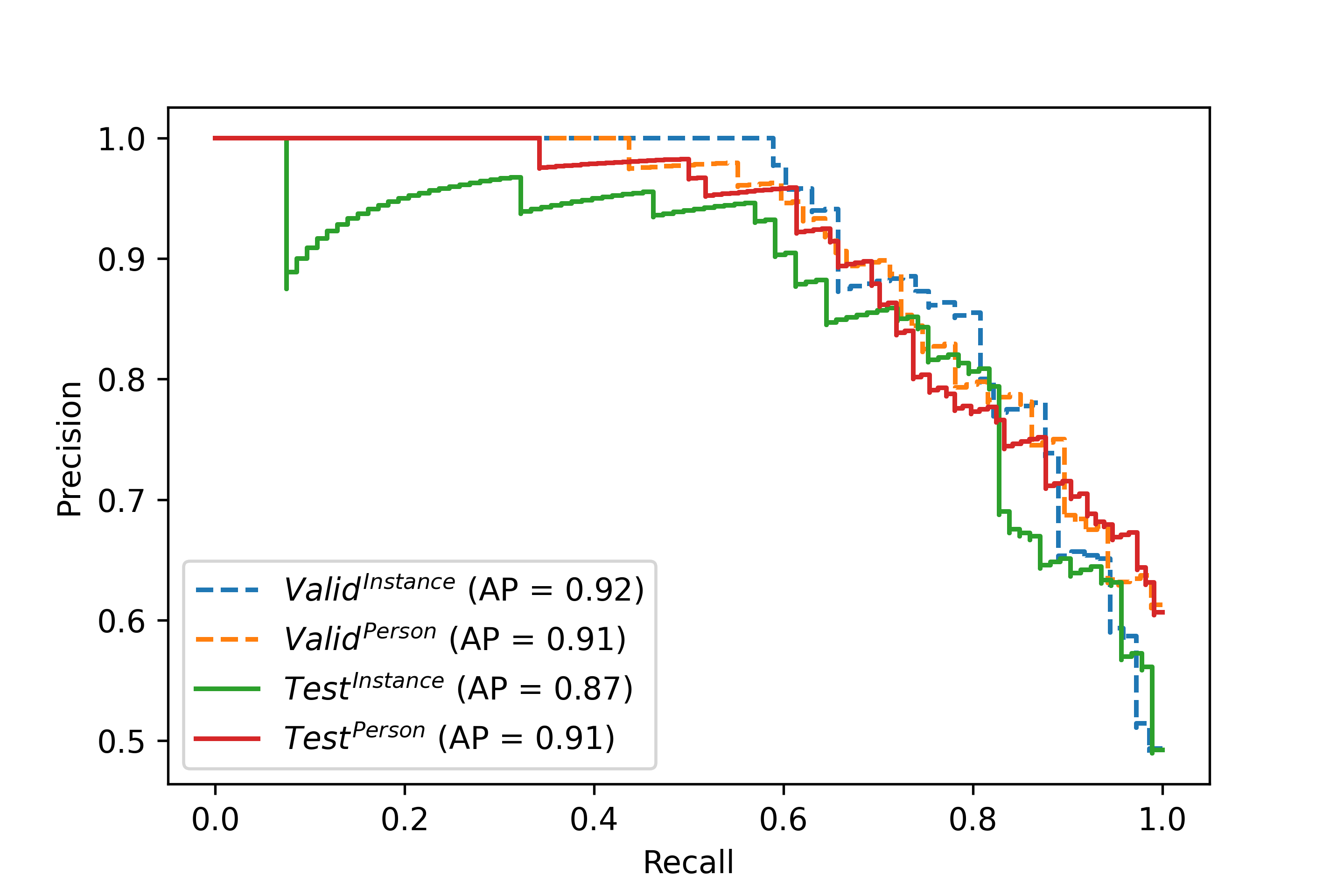}
    \caption{PR curve (SVM on labeling video recordings): the dash lines correspond to the validation set while the solid lines are for the test set. Average precision scores are also provided in the legend as AP. While the blue and green are for the dataset with instance labeling, the orange and red indicate values for the dataset with person labeling. }
    \label{fig:PR}
\end{figure}

For the third part, patient membership inference, the performance of the SVM has been illustrated in Fig. \ref{fig:ROC} and \ref{fig:PR} where the receiver operating characteristic (ROC) curve and precision-recall (PR) curve and trapezoidal area under the curve (AUC) and average precision (AP) have been shown for both validation and test sets with instance and person labels. The accuracy and F1-score are shown in table \ref{table:accuracy_F1}.

Table \ref{table:Numbers} summarises the information when considering only the records where the instance and person labels are different.

\small\begin{table}[htbp]
\caption{Validation loss (binary cross-entropy) scores for different inputs for the frame labeling network, for instance (inst), and person (Prsn) labeling$^{\mathrm{a}}$.}
\label{table:loss_param}
\begin{center}
\begin{tabular}{l|l|l|l|l|l|}
\cline{2-6}
 &
  2 output$^{\mathrm{b}}$ &
  +2 grad$^{\mathrm{c}}$ &
  +5 grad$^{\mathrm{d}}$ &
  \begin{tabular}[c]{@{}l@{}}+2 grad\\ + loss\\ + label$^{\mathrm{e}}$\end{tabular} &
  \begin{tabular}[c]{@{}l@{}}+3 grad\\ + loss\\ + label$^{\mathrm{f}}$\end{tabular} \\ \hline
\multicolumn{1}{|l|}{loss$_{Inst}$} &
  0.7 &
  0.61 &
  0.61 &
  0.57 &
  0.57 \\ \hline
\multicolumn{1}{|l|}{loss$_{Prsn}$} &
  0.7 &
  0.61 &
  0.62 &
  0.59 &
  0.58 \\ \hline
\end{tabular}
\begin{tablenotes}
\item[1] $^{\mathrm{a}}$For reference, the baseline would be $- log(1/2) = 0.693$.
\item[2] $^{\mathrm{b}}$Takes the two last outputs (the output and the layer just before it) of the target model as the input.
\item[3] $^{\mathrm{c}}$Takes the two last outputs and also the two gradients before the last gradient of the target model. 
\item[4] $^{\mathrm{d}}$Takes all the "+2 grad" and also the last gradient of three different sections of the target model (boundary, face, eyes).
\item[5] $^{\mathrm{e}}$Takes the two last outputs and also the two before the last gradient and label and loss of target model.
\item[6] $^{\mathrm{f}}$Takes the two last outputs and also the three last gradients and label and loss of target model.
\end{tablenotes}
\end{center}
\end{table}

\small\begin{table}[htbp]
\caption{Accuracy (Acc) and F1-score (F1) for SVM in validation and test data with both instance and person label sets$^{\mathrm{a}}$.} 
\label{table:accuracy_F1}
\begin{center}
\begin{tabular}{l|l|l|l|l|}
\cline{2-5}
         & Valid$_{instance}$ & Valid$_{person}$ & Test$_{instance}$ & Test$_{person}$ \\ \hline
\multicolumn{1}{|l|}{Acc} & 0.85            & 0.81         & 0.79           & 0.77         \\ \hline
\multicolumn{1}{|l|}{F1} & 0.82            & 0.80          & 0.80           & 0.79         \\ \hline
\end{tabular}
\begin{tablenotes}
\item[1] $^{\mathrm{a}}$Threshold have been adjusted to achieve the best performance on the validation set in each model (0.68 for instance and 0.8 for person model).
\end{tablenotes}
\end{center}
\end{table}

\small\begin{table}[htbp]
\caption{Performance for people with multiple recordings}
\label{table:Numbers}
\begin{center}
\begin{tabular}{l|l|l|l|l|}
\cline{2-4}
                & Train & Valid & Test \\ \hline
\multicolumn{1}{|l|}{Total$^{\mathrm{a}}$}           & 19    & 14    & 21   \\ \hline
\multicolumn{1}{|l|}{Instance$^{model}$ $^{\mathrm{b}}$} & 11    & 4     & 11   \\ \hline
\multicolumn{1}{|l|}{Person$^{model}$ $^{\mathrm{c}}$}   & 16    & 8     & 12   \\ \hline
\multicolumn{1}{|l|}{$1-$P-value$^{Person}$ $^{\mathrm{d}}$}   & 0.996    & 0.21     & 0.44  \\ \hline
\end{tabular}
\end{center}
\begin{tablenotes}
\item[1] $^{\mathrm{a}}$The total number of recordings in each set that belongs to a person who has another recording present in the training set of the target model (eye-tracking model).
\item[2] $^{\mathrm{b}}$Only provided for the sake of completeness and is the number of picked recordings as inside, despite being trained on them with labeling as outside.
\item[3] $^{\mathrm{c}}$The number of these recordings that had been picked as inside in model trained on person labels.
\item[4] $^{\mathrm{d}}$For when the probability of success in a Bernoulli experiment is 50$\%$ is provided to illustrate the significance of numbers.
\end{tablenotes}
\end{table}

\section{Discussion}
The attempt of the three-stage pipeline to attack the target model is to investigate the potential memorization of participants' facial information. One should note that here the labels of the target model are the gaze coordinates of participants. What makes it much more challenging to harvest information from, compared to a typical classification model, for example (as used in ref\cite{nasr2019comprehensive}), is that the facial properties of a person are not correlated to the labels in a general manner. For instance, two people with very different facial features can look at exactly the same spot. 

The pipeline design has been adopted because the target model is not a simple network and since there are different lengths of frames for recording. Also, this design reduces the number of learning parameters of the membership inference model. So the Patient membership inference and frame classifier were not trained together in an end-to-end manner. Table \ref{table:loss_param} shows that the use of only outputs of the target model for the frame labeling part of the attack gives no advantage over a random classifier. This indicates that a simple black-box attack that only uses outputs is likely to fail. 

As can be seen from table \ref{table:loss_param} adding different gradients and also labels and loss of the target model make the classification network much more useful to decipher the correct labels both for the person and instance labels/models. The model with the highest performance on the validation set was selected for this part of the pipeline. These labels for each frame were taken as the input for the next step, where a linear SVM was selected as it gave the highest values for AP score and AUC-ROC. The values in table \ref{table:accuracy_F1} and also Fig. \ref{fig:ROC} and \ref{fig:PR} show the performance of this third part of the pipeline. They show that the labeling is reasonably successful in determining if a recording has been part of the target model training set or not. 

From these results, one may assume the participants' information is recorded in the trained network and can be extracted successfully to identify them. But this does not translate to the identifiable information leak. Firstly, for this attack, the assumption is that the attackers not only have full access to the target model but also access to a third of the recordings with the knowledge that they have been in the training set of the target model. Moreover, they have access to the other two-third of the training recordings, and the only hassle is that those are mixed with a similar number of recordings which has not been in the training set, and they only have to infer which half of those are the ones in the training set. 

Indeed the significance of the two labels used can be appreciated here since the difference between the two labels is the difference of the exact recording vs. the same person (and so, same face) but in a different recording. So to see if these recorded data are facial specifications of the participants recorded in the network, or more like the properties of the specific frame used in the training set, we investigate the ones with more than one recordings. Table \ref{table:Numbers} shows that from 54 of such recordings, 19 had been used in attack model's training set while 14 are in the validation set and 21 in the test set. These are the recordings that are not used in the training of the target model directly but are from the people who have other recordings present in that training set. Suppose the facial features of participants recorded in the network are making the predictions of the first part possible. In that case, they should show their effectiveness in the "Person" model (model trained by Person Y labels) in labeling these participants since they are from the same people but only different recordings. The numbers in table \ref{table:Numbers} do not support this claim. While the training set gets 16 out of 19, which in terms of significance produce the $1-$P-value$=0.996$, the validation and test sets get 8 out of 14 and 12 out of 21, which is completely close to random and not significant by any means. This suggests that the high performance of the pipeline in differentiating between the in and out of target training set does not come from the facial features of the participants but other aspects and features of the specific frames in the set. Although one limitation is that the number of training samples here is much smaller than a single recording dataset and can potentially be changed with more data, the argument that more data can change the results can always be raised in any specific data-dependent analysis.

While techniques like differential privacy (DP) can guarantee mathematically provable privacy preservation and robustness against many attacks\cite{dwork2014algorithmic}, they have other limitations. The implementation of DP is not a trivial task in different platforms, especially in complex algorithms, and if not correctly implemented, it can be deceiving with a false sense of security\cite{mironov2012significance}. Also, the performance of the models can suffer drastically by the implementation of DP\cite{rahman2018membership}\cite{friedman2010data}, especially when the size of the training set is limited. Therefore, in the real world, the availability and portability of the data are also critical. While one needs to take all the measures to protect sensitive or private data, it is also essential to be aware that no golden bullet is present to implement in every context.

\section{Conclusion}
While the proposed pipeline exhibits good performance for differentiating between  recordings taken from different individuals, an analysis with multiple recordings captured from given individuals demonstrates that the performance of a classifier drops to the level of a random guess when attempting to identify whether an individual appeared in the training set. This provides strong evidence that it is unlikely for recognizable facial features to be recorded in the target model.  While this is not an exhaustive evaluation, and doesn't preclude the possibility that future research will find a way to identify individuals within the network, we have found no current method to do so at this time.

\section{Acknowledgment}
GC and SS are funded by the National Science Foundation, grant number 1822378 ‘Leveraging Heterogeneous Data Across International Borders in a Privacy Preserving Manner for Clinical Deep Learning’. GC is partially supported by the National Center for Advancing Translational Sciences of the National Institutes of Health under Award Number UL1TR002378. The content is solely the responsibility of the authors and does not necessarily represent the official views of the National Institutes of Health. GC an AL are inventors of the eye-tracking algorithm described in this study, which has been licensed out to Linus Health.
 
\bibliographystyle{IEEEtran}
\bibliography{SalRef} 

\clearpage
\end{document}